\documentclass[conference]{IEEEtran}
\IEEEoverridecommandlockouts
\usepackage{color}
\usepackage{cite}
\usepackage{amsmath,amssymb,amsfonts}
\usepackage{algorithmic}
\usepackage{graphicx}
\usepackage{textcomp}
\usepackage{xcolor}
\def\BibTeX{{\rm B\kern-.05em{\sc i\kern-.025em b}\kern-.08em
    T\kern-.1667em\lower.7ex\hbox{E}\kern-.125emX}}
    
\usepackage[utf8]{inputenc}
\usepackage[]{lipsum}
\usepackage[]{tabulary}
\usepackage{caption}
\usepackage{subcaption}
\usepackage{amsmath}
\usepackage{wrapfig}
\usepackage{tikz}
\usetikzlibrary{bayesnet}
\usepackage{multirow}
\usepackage[normalem]{ulem}

\begin{document}

\title{SummerTime: Variable-length Time Series Summarization with Applications to Physical Activity Analysis}


\author{
\IEEEauthorblockN{Kevin Amaral}
\IEEEauthorblockA{\textit{Computer Science}\\
\textit{University of Massachusetts Boston}\\
Boston, MA \\
Kevin.M.Amaral@gmail.com}\\
\IEEEauthorblockN{Scott Crouter PhD.}
\IEEEauthorblockA{\textit{Exercise and Health}\\
\textit{University of Tennessee-Knoxville}\\
Knoxville, TN \\
scrouter@utk.edu}
\and
\IEEEauthorblockN{Zihan Li}
\IEEEauthorblockA{\textit{Computer Science}\\
\textit{University of Massachusetts Boston}\\
Boston, MA \\
Zihan.Li001@umb.edu}\\
\IEEEauthorblockN{Ping Chen PhD.}
\IEEEauthorblockA{\textit{Computer Engineering}\\
\textit{University of Massachusetts Boston}\\
Boston, MA \\
Ping.Chen@umb.edu}
\and
\IEEEauthorblockN{Wei Ding PhD.}
\IEEEauthorblockA{\textit{Computer Science}\\
\textit{University of Massachusetts Boston}\\
Boston, MA \\
wei.ding@umb.edu}
}







\maketitle
\begin{abstract}
\textit{SummerTime} seeks to summarize globally time series signals and provides a fixed-length, robust summarization of the variable-length time series. Many classical machine learning methods for classification and regression depend on data instances with a fixed number of features. As a result, those methods cannot be directly applied to variable-length time series data. One common approach is to perform classification over a sliding window on the data and aggregate the decisions made at local sections of the time series in some way, through majority voting for classification or averaging for regression. The downside to this approach is that minority local information is lost in the voting process and averaging assumes that each time series measurement is equal in significance. Also, since time series can be of varying length, the quality of votes and averages could vary greatly in cases where there is a close voting tie or bimodal distribution of regression domain. Summarization conducted by the \textit{SummerTime} method will be a fixed-length feature vector which can be used in-place of the time series dataset for use with classical machine learning methods. We use Gaussian Mixture models (GMM) over small same-length disjoint windows in the time series to group local data into clusters. The time series' rate of membership for each cluster will be a feature in the summarization. By making use of variational methods, the GMM converges to a more robust mixture, meaning the clusters are more resistant to noise and overfitting. Further, the model is naturally capable of converging to an appropriate cluster count. We validate our method on our [The dataset is created using our NIH awards. The name is removed in compliance with Triple Blind] dataset, an imbalanced physical activity dataset with variable-length time series structure. We compare our results to state-of-the-art studies in physical activity classification and show high-quality improvement by classifying with only the summarization. Finally, we show that regression using the summarization can augment energy expenditure estimation, producing more robust and precise results.
\end{abstract}


\begin{IEEEkeywords}
Time Series, Clustering, Classification, Regression, Summarization
\end{IEEEkeywords}

\section{Introduction}
The physical activity (PA) is one of the most important areas when people try to know more about themselves.To record the activities, time series data is abundant in the PA, especially when researchers record the activities in the dynamic way. In this paper, we propose a data-driven based method which provides a variable-length summarization of physical activity tracks.

The physical activity is defined as any movement of the body that requires energy expenditure. Since the activities always happen with multiple body movements, we named a sequence of single movements as a specific activity. Then, the different combinations of single breakdown movements are defined as different PA. Researchers need to match the real movement track with the pre-defined activities which contains a specific sequence of movements. Since the different PAs may consist of the same sub-sequence of single body movements, it is crucial to decide how to break the full movement track. However, it is hard to set an appropriate break in the movement track based on the traditional PA technique. In this paper, we apply the data-driven machine learning method to provide a way to redefine the PAs as different clusters. These clusters will have significant common patterns but may not follow the traditional PA definitions (like the specific sequence of single movements). With the data-driven model, the data can reveal their hidden patterns by itself. Since no pre-defined movement sequence is required, the chance bias can be reduced significantly. Another benefit of our data-driven method is to improve the accuracy by considering the small effective events into the model. The small effective events can also have significant effects to the final result, like the long-tail effect. The information from both majority and minority of effective events will contribute in the method equally.


Our method, \textit{SummerTime} (\textbf{Time} series \textbf{Summ\sout{a}$ ^\text{e}$r}ization), leverages the features of time windows as local information about the time series signal and use them to create a global description of the signal which preserves the information of every time window within it. The traditional time series machine learning methods depend on a fixed-length representation so that they can build relationships between features and use those to produce their results. However, it is difficult to apply classical time series machine learning methods to the time series PA data. Since, though, the time series PA data is rich in information, it is rarely uniformly structured across instances, differing in length and scale. For instance, in regression methods, the time series data can be trained by treating the shorter length of features. However, building models in this way has to loss parts of information in the data, which would lead to overfitting and to reduce the accuracy of models working with future unseen longer features. Our model, \textit{SummerTime}, attempts to encapsulate all global information about the time series PA signal as a fixed-length representation, producing a feature vector with a constant number of features for each instance. Not only including as much as information from the data, but also providing a competitive solution to deal with variable-length time series PA data. The features of this global description are learnt latently using clustering of the local time windows. This effectively treats the bias from manually-constructed features and results in a higher-quality interpretation of the features. This does not, however, fully remedy the bias from the manually constructed time window features. The prime assumptions we make is that the time window features are already of moderate-to-significant quality for locally describing the signals and that the main obstacles of other time series methods are in the additional assumptions they undertake to solve their respective problems. These assumptions have merit as we are able to generate from the local time window features potent summary features, and we observe many opportunities for competitive approaches to introduce biases and make our best efforts to avoid those biases. In accounting for these biases, we observe overall better performance than each of our competitive methods. We hope to resolve any further biases by having an end-to-end data-driven solution. Our contributions are as follows:

\begin{itemize}
    \item \textbf{Fixed-length feature-vector representation of variable-length time series:} This is one of the major obstacles that variable-length time series faces. Many approaches seek to instead classify and regress on local information about the signal or to resize the signal to uniform length across all instances. In either case, these approaches usually achieve the fixed-length representation so as to use classical methods. However, rescaling the signal relinquishes information about the proportion to time (i.e. the concept of an $n$-second interval is lost) and voting over the classifications and regressions done in the local time windows fails to encode global information about the signal as a whole. Our model achieves the representation without losing proportional and global information.
    \item \textbf{End-to-end data-driven approach:} Our method, \textit{SummerTime}, is designed to fit between the time series dataset and the classical methods you want to apply, performing feature construction independently of the experimenter and extracting all knowledge from the data itself. While this fits the idea of a black box, we are of the opinion that this is a valuable trait in a framework. By having the method be data-driven, we avoid the opportunity for introducing further bias from the experimenter and reduce the loss of information.
    \item \textbf{Demonstration of improved robustness using only summarization features:} With our dataset, we show that classification using one of the other schemes, specifically voting over local time windows, is vulnerable to cross-class classification error between far physical activity types. We then show that using only features extracted by the \textit{SummerTime} method, it was able to out-perform them overall, and also significantly improved the misclassification between near classes, which is a major demonstration of robustness.
    \item \textbf{Demonstration of improved precision using summarization features to augment regression:} Again, in our dataset, we show that we can augment existing state-of-the-art regression models by including our \textit{SummerTime} features. Not only does doing so improve regression results, but also note that the regression RMSE of the Run-type activities, a close class to the Walk-type activities, drops to around the same level as Walk-type activities. This increased precision indicates that the \textit{SummerTime} features overcome some bias clearly present in all other regression methods we compare against, each with run-type error well-above Walk-type error.
\end{itemize}

The rest of this paper is organized as follows: The second section introduces the time series PA dataset we are going to analysis. The third section goes on to describe the problem of variable-time series and the goal of classification and regression on those types of data. In the forth section, we describe our proposed data-driven method, the \textit{SummerTime} Framework, which solve the problem outlined in the third section of the paper. In the fifth section, we present our experiment results, and the interpretations thereof. In the sixth section, we present related work. Finally, we conclude in the seventh section.

\begin{figure*}
    \centering
    \includegraphics[width=\linewidth,angle=90]{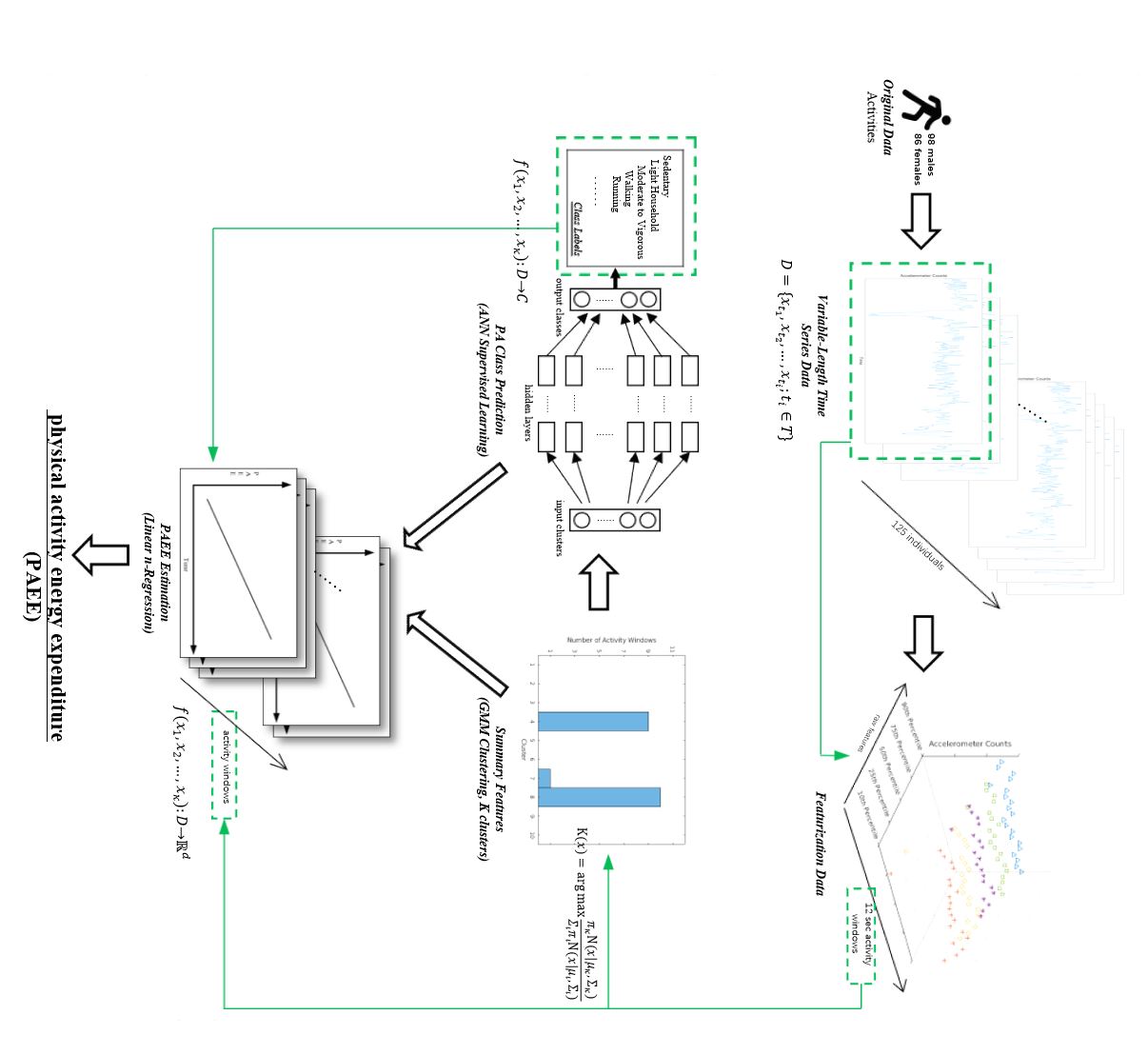}
    \caption{The \textit{SummerTime} framework diagram. The local window features are clustered to produce new summary features for use in the Classification and Regression phases. The Classification Phase makes use of only the newly constructed summaries from the Clustering Phase. In our experimental setting, the classification is of physical activity type. In the Regression Phase, the Window Features, the Summary Features, and the Classification result are all used to produce the physical activity energy expenditure estimate.}
    \label{fig:frameworkfigure}
\end{figure*}

\section{Data Description}
\subsection{Dataset}

In the medical field, measuring physical activity in children is crucial in learning the link between physical activity and health problems such as obesity and other metabolic issues. Measuring physical activity is generally done in two ways: (1) identifying the types of activities performed and (2) measuring caloric or energy expenditure. Accelerometers prove to be the ideal device for physical activity data collection. While they do not directly identify activities being performed nor measure caloric expenditure, they provide an objective measurement which is information-rich and can be analyzed to predict and estimate these two measurements. They are also a low-cost and non-invasive alternative to indirect calorimetry, which measures caloric expenditure through a breathing device. However, analyzing accelerometer data to make these predictions and estimates is still a difficult, unsolved problem and estimates made with state-of-the-art methods are still high-variance \cite{freedson2011evaluation}. Furthermore, classical machine learning methods cannot be applied out-of-the-box to a free-living setting since physical activity accelerometer data collected outside of a laboratory environment is generally unstructured or continuous stream time series.

The dataset used for validation is a combined dataset from two studies performed by [Citation and Name Removed in compliance with Triple Blind]. The data was collected from child participants performing a wide variety of physical activities. The participants were outfitted with an Actigraph X accelerometer which was used to collect tri-axial accelerometer counts at one count per second. Participants were hooked up with a Cosmed K4$\text{b}^2$ to collect energy expenditure measurements in MET (Metabolic Equivalent of Task) units. 

Start and end times were observed, as well as physical activity (PA) type for every bout of activity performed by each participant. Our model will not predict cut-points for activities but will instead classify the PA type and estimate physical activity energy expenditure (PAEE) of time series segments. As such, our dataset comes pre-segmented with true segment bounds.

The variable-length nature of the data is an allegory for the free-living setting, where activity length is not rigidly defined. The common approach to deal with multi-instance data, data for which each instance has multiple parts, is break it up into its smaller parts and build a model over the members of the multi-instances instead. We will do just that, but we will maintain the association between the multi-instance and its members by referring to the multi-instance as the activity bout and its members as activity windows which belong to the bout.

In total, 184 child participants' data were used. There were 98 male participants between 8 years old and 15 years old. There were 86 female participants between 8 years old and 14 years old. 

For each of these participants, one bout of lying resting for up to 30 minutes with a median time of 17 minutes. All other activities were performed for up to 10 minutes with a median time of 4 minutes. All activity date is collected at 1-second resolution. Activities which were included in the study were Computer Games, Reading, Light Cleaning, Sweeping, Brisk Track Walking, Slow Track Walking, Track Running, Walking Course, Playing Catch, Wall Ball, and Workout Video. These activities classes were binned into categories by the rigorousness of the associated activity; the category classes were Sedentary (Sed), Light Household Chores and Games (LHH), Moderate to Vigorous Household Chores and Sports (MtV), Walking (Walk), and Running (Run). See Table \ref{tab:activities} for a visual breakdown of the categories. The categories are organized in order of ascending intensity.

The representation of each activity category in the dataset is not equal. Reference again Table \ref{tab:activities} to see the huge disparity in the number of windows.

\section{Problem Description}
Physical activities (PA) of human consist of the elemental movements of body, such as leg raising and arm stretching. We define each breakdown of movement as an atomic event of the time series activities data. And each observed single activity is performed as a PA signal. In this case, each activity signal is the combination of multiple activity events. The time series PA data are a discrete sequence of events of the form $D = \{x_{t_1}, x_{t_2}, \dots ; t_i \in T\}$ such that their index set $T$ is temporal in nature. In many cases, the events do not need to occur at a constant frequency but for the purposes of this paper, we will consider only the case when the time series has events at a constant frequency. 

The most popular two techniques of prediction methods are classification and regression. However, both of them are not suitable for the variable length time series PA activities data. The reasons and solutions are introduced as follow.

Classification is the process in which a trained model maps time series data to a nominal class label. 
Traditional classification methods operate on the fixed number of features, such that they can be generalized as follows, where $\mathcal D$ is the space of instances which have $k$ features and $\mathcal C$ is the space of class labels.

\begin{equation}
    f(x_1, \dots, x_k) : \mathcal D \to \mathcal C
\end{equation}

For a fixed-length vectors dataset, we can apply those traditional methods out of the box. However, in the case of variable-length time series, it is impossible to apply traditional models directly.

One current solution to this is to break-up each time series signal into equally-sized time series windows and to perform traditional classification over features of the windows. The final result would be aggregated through voting or taking the mode. This results in loss of minority information within the signal. However, minority information may also contribute to the classification, especially when the difference is not very significant. For example, in the cases where a signal has equal parts of two classes, the method observes 50\% confidence in its classification result. This is akin to a coin flip.

Another current solution to the variable-length time series data is to rescale the time series signal so that all instances are of the same length. If done through interpolation, estimating and including in-between points, we introduce bias from our choice of interpolation method. In non-smooth signal domains, estimates of in-between points can be radically inaccurate. Regardless of if we account for that bias somehow, rescaling the time series removes proportionality from the signal, relinquishing the distinction between an $n$-second interval and an $m$-second interval. This causes another bias in that $n$-point intervals are treated equally, even if they correspond to vastly different stretches of real time.

In this paper, instead of convert variable-length time series data into a fixed-length summarized format, we propose a method to do this in a way that preserves global information about the signal (i.e. no data point's contribution is unaccounted for) and preserves in someway proportionality. Further more, our method works in a data-driven manner to avoid biases causing by human being, such as the experiments or the labeling process. 


Our solution is to cluster the time series window features and then use the ratios of membership to each cluster as the features of the summary. This technique can fully encapsulate the global information of the PA signals since the inclusion or exclusion of any events in the signal will affect every summary feature. In this way, the proportionality can be accounted by comparing the variable length time series signals without treating their events as equally frequent. Finally, the unsupervised clustering technique is applied to construct the summary of the signals, which can avoid the bias causing by humans when assigning signals into the targeted clusters significantly. 



Regression part has a similar process to classification in that we are building a model which maps from a time series data space to a target space. However, the target space of regression methods is generally numeric and continuous, rather than categorical and discrete as in a nominal space of classification. Regression models are usually of the form:

\begin{equation}
    f(x_1, \dots, x_k) : \mathcal D \to \mathbb R^d
\end{equation}

The same as classification case, traditional regression models only can operate over data with the same number of components. One current solution for applying regression models to variable-length data is to treat shorter instances as having missing values. The problem of this is that there are variables which become specifically tuned to long instances from the training set. When presented with new long-instance test data from a different domain distribution, those variables will impact the regression results in ways the training process cannot account for.

Another current solution is to perform the regression locally on windows and aggregate the regression result as a total or average or similar statistic over the windows. In cases where the distribution of windows is multi-modal, an equally-weighted average can push results in the direction of the majority, heavily influenced by the number of points in the signal. This is not ideal since it leads to the issues of proportionality like it in the classification. However, this influence can be lessened without having to make major changes to the underlying regression model.

Our solution to the regression problem is to apply multiple distributions following the clusters from the classification results. In cases where the PA signals to be regressed over have separate classifications, are different from each other in some categorical way, regression models can be built for each category to produce better results than if a single model was designed to handle every category. So, to decide on which regression model should be applied to a new time series PA signal, the classification of its events is necessary. The regression model would be applied to each window and aggregate the result over the whole signals as usual. But, we will include, as regression variables, the time series summarization of the signal from the clustering. These variables carry information about the signal as a whole which we demonstrate to have a positive effect on time series of differing lengths.

\begin{figure*}
    \centering
    \begin{subfigure}{\textwidth}
    \centering
        \includegraphics[width=0.8\linewidth]{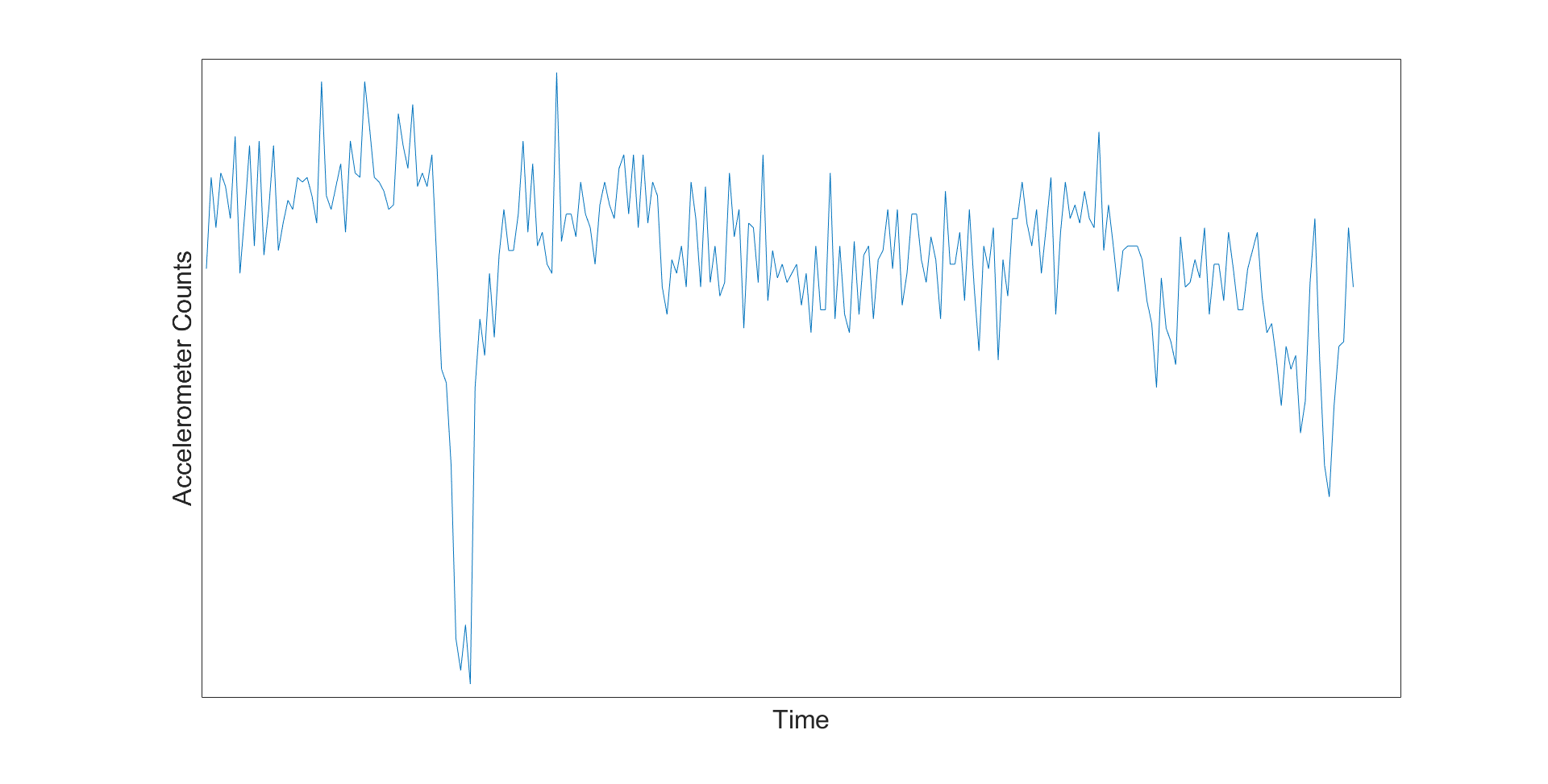}
        \label{fig:zero}
        \caption{}
    \end{subfigure}
    \begin{subfigure}{0.5\textwidth}
    \centering
        \includegraphics[width=\linewidth]{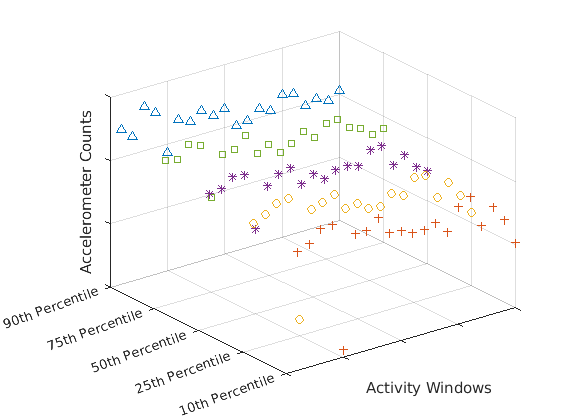}
        \label{fig:one}
        \caption{}
    \end{subfigure}%
    \begin{subfigure}{0.5\textwidth}
    \centering
        \includegraphics[width=\linewidth]{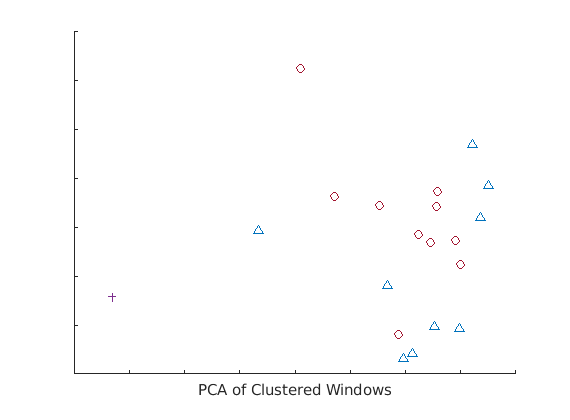}
        \label{fig:two}
        \caption{}
    \end{subfigure}
    \begin{subfigure}{0.5\textwidth}
    \centering
        \includegraphics[width=\linewidth]{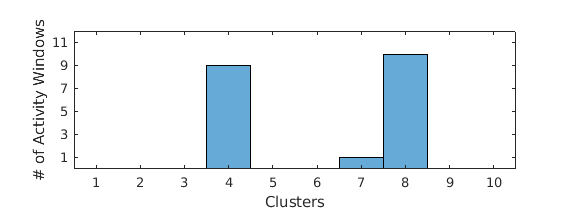}
        \label{fig:three}
        \caption{}
    \end{subfigure}%
    \begin{subfigure}{0.5\textwidth}
    \centering
        \includegraphics[width=\linewidth]{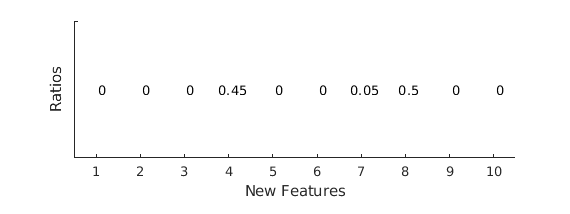}
        \label{fig:three2}
        \caption{} 
    \end{subfigure}
    \caption{Above, we show a single instance of Track Running as it makes its way through the clustering phase of the framework. \textbf{(a)} shows the x-axis accelerometer counts for the activity. \textbf{(b)} shows the breakdown of the signal into the 12-second windows for this axis. Not shown is the lag-1 autocorrelation feature. This is the standard means for constructing data instances from physical activity timeseries data. In \textbf{(c)}, we show a PCA projection of the entire activity. The shapes of the data points indicate the clusters each activity window was assigned to. \textbf{(d) and (e)} demonstrate the feature construction from the clustering. The new features associated with an activity instance are the ratios of membership to each cluster.}

    \label{fig:figure1}
\end{figure*}

\section{Methodology}
\subsection{Modeling Approach}
Compared with the traditional methods, our SummerTime algorithm works on both the fixed-length and variable-length time series data. Further more, our model can provide very competitive results of regression technique by considering multiple distributions for different kinds of PA signals. 

There are three major phases in the algorithm. The \textit{SummerTime} framework can be introduced as follow:
\begin{enumerate}
    \item Clusters Generation: Produce Time Series Summarizations
    \item Classification: Determine the Class of the Time Series Signals from the Summarization
    \item Regression: Select the Appropriate Regression Model Based on the Previous phases and Leverage Summaries to Refine Regression over the Time Series Signals
\end{enumerate}

\subsubsection{Cluster Generation Phase}
In this stage, we need to extract the summarization of clusters from the signals with different events. 

Gaussian Mixture Models (GMM) are a probability distribution consisting of a linear combination of multivariate Gaussian distributions each with their own mean and variance \cite{bishop2006pattern}. These individual Gaussian distributions will each correspond to a cluster of time-series activity windows which are most likely to belong to them. We utilize this cluster phase of the framework to produce a feature vector of ratios for each time series signal. This feature vector's components are of the form $\frac{N^{n}_k}{N^{n}}$ where $N^{n}_k$ is the number of windows $n$ that belong to cluster $k$ and $N^{n}$ is the total number of windows in the signal $n$.

Using the features of the time series windows, we can define the distribution $p(x)$ over all windows $x \in X$. It is reasonable to assume that within the distribution, there are $K$ component distributions to which each point $x$ truly belongs. We further assume that those components are all normally distributed. As a result of these assumptions, we have Equation \ref{eq:distribution} as the distribution function.

\begin{equation}
p(x |\pi, \mu, \Sigma) = \sum_{k=1}^K \pi_k \mathcal N(x|\mu_k, \Sigma_k)
\label{eq:distribution}
\end{equation}

Here, $\pi_k$ is referred to as a mixing coefficient of the $k$-th component of the mixture distribution. We take $\sum \pi_k = 1$ so that $\pi_k$ acts as the probability of an arbitrary $x$ belonging to component $k$.

For $x$ in the distribution, we can measure the probability of $x$ belonging to component $k$ by taking Equation \ref{eq:responsibility}. This is referred to as $\gamma_k(x)$, the responsibility of component $k$ for $x$.

\begin{equation}
\gamma_k(x) = \frac{\pi_k \mathcal N(x|\mu_k, \Sigma_k)}{\sum_l \pi_l \mathcal N(x|\mu_l, \Sigma_l)}
\label{eq:responsibility}
\end{equation}

To determine which component $k$ a data point $x$ belongs to, we can take the maximum of the responsibilities $\gamma_k(x)$ over $k$.

\begin{equation}
\mathcal K(x) = \mathop{\arg \max}_k \gamma_k(x)
\end{equation}

We learn this model using variational Bayesian techniques which allow us to learn an accurate number of clusters $K$ as well as learn each of the $K$ clusters very accurately without overfitting or producing singularity clusters \cite{nasios2006variational}.

This clustering function $\mathcal K(x)$ allows us to create the fixed-length feature vectors for the dataset. We'll now define $N_k^n$ as follows where $I_k$ is the usual indicator function and $\mathcal X_n$ is the set of all activity windows belonging to activity $n$.

\begin{equation}
N_k^n = \sum_{x}^{|\mathcal X_n|} I_k(\mathcal K(x))
\end{equation}

\begin{equation}
N^n = |\mathcal X_n|
\end{equation}

This is the end-point for the clustering phase of the framework. For each activity $n$, we now have a single fixed-length vector $\left[\frac{N^{n}_1}{N^{n}}, \dots, \frac{N^{n}_K}{N^{n}}\right]^T$.

\subsubsection{Classification Phase}

We choose Artificial Neural Network (ANN) models for classification. An ANN is a layered graph-based non-linear regression model commonly used in machine learning and statistical settings \cite{bishop2006pattern5}. There is precedent in applying ANNs for classification of time series signals \cite{Staudenmayer2009} \cite{Trost}. In this phase of the \textit{SummerTime} framework, we are going to perform classification given only the feature vectors produced in the previous clustering phase.

As a result, our ANN model will consist of $K$ input nodes, where $K$ is the latently-learned cluster-count variable corresponding to the number of clusters. The hidden layer consists of 25 hidden neurons. The output layer is a softmax layer which produces discrete categorical values corresponding to class predictions.

One drawback to using an ANN is that the model is generally considered a blackbox and is often over-parameterized. As a result, the number of hidden layer nodes was chosen empirically. However, this does not inhibit the model's strength at performing classification. By selecting empirically the hidden layer hyperparameter, we can assume that the network with the appropriate number of hidden layer neurons will perform best.

This is the end-point for the classification phase of the framework which produces our final classification prediction for each time series signal. We then feed-forward the fixed-length feature vector from the clustering phase and this classification result forward into the regression phase.

\subsubsection{Linear $n$-Regression Phase}

Results from the Two-Regression Model by \cite{Crouter1324} show that knowledge of a time series signal's class has a significant impact on accurate regression over the signal. We intend to leverage not only the physical activity class prediction of the classification phase of the framework, but also to reuse the fixed-length feature vector produced in the clustering phase of the framework.

For the regression model, we use $|\mathcal C|$ distinct linear regression models. Each model is trained independently of the others, each on one of the time series classes. Using the classification prediction, we select which of the  $|\mathcal C|$ to use for the regression phase. The regression model will be applied over each window, rather than the full-length signal.

Each model is a linear regression model of the following form:

\begin{equation}
y = X\beta + \varepsilon
\end{equation}

The rows of our design matrix $X$ correspond to time series windows. The columns of our design matrix $X$ consist of a bias term $1$, the features of time series window, and finally it also includes the $K$ additional cluster features constructed for the full-length time series signal. In total, the number of variables over which the regression is performed is $1+W_\text{count}+K$, where $W_\text{count}$ is the number of features in the window. The model $\beta$ is to be of the appropriate size and will be learned with minimal error $\varepsilon$. The regression target is represented by $y$.

\begin{table*}
    \centering
    \begin{tabular}{|c|c|c|c|}
            \hline
            \multicolumn{2}{|c|}{Category and Class} & \multicolumn{2}{|c|}{Number of Windows} \\ \hline
            \multirow{ 3}{*}{Sedentary} & Lying Rest & 14755 & \multirow{ 3}{*}{16475}\\ \cline{2-3}
            & Playing Computer Games & 860 & \\ \cline{2-3}
            & Reading  & 860 & \\ \hline
            \multirow{ 3}{*}{Light Household and Games} & Light Cleaning & 840 & \multirow{ 3}{*}{2505}\\ \cline{2-3}
            & Sweeping  & 865 & \\ \cline{2-3}
            & Workout Video  & 800 & \\ \hline
            \multirow{ 2}{*}{Moderate-Vigorous Household and Sports}& Wall Ball & 845 & \multirow{ 2}{*}{1570}\\ \cline{2-3}
            & Playing Catch  & 725 & \\ \hline
            \multirow{ 3}{*}{Walk} & Brisk Track Walking & 1210 & \multirow{ 3}{*}{3775} \\ \cline{2-3}
            & Slow Track Walking  & 1000 & \\ \cline{2-3}
            & Walking Course  & 1565 & \\ \hline
            \multirow{ 1}{*}{Run} & Track Running & 485 & \multirow{ 1}{*}{485} \\ \hline
    \end{tabular}
    \caption{Here we show the types of activities present in the data set and what activity categories they fall under. On the right side, the columns are the number of 12-second windows corresponding to each category and class. As you can see, Sedentary activities are significantly over-represented within the data set and Running activities are under-represented, contributing to a severely imbalanced data set. Furthermore, Walking is the second most-represented category, having nearly eight times as many windows as Running, a class that it is difficult to be distinguished from.}
    \label{tab:activities}
\end{table*}

\section{Experiments and Results} 
\subsection{Window Features}

For use with our method, a time series signal must be segmented into equally-sized intervals or windows. Each window must be assigned features that characterize the local information in that region of the signal. For this purpose, we chose to use percentile information, which has precedence in other works \cite{Staudenmayer2009} \cite{Trost} .

For percentile features, we characterize a window along each of its axes by the interval's 10th, 25th, 50th, 75th, and 90th percentiles. This follows the scheme used by those previous works. This encodes information about the distribution of points within the window, excluding the minimum and maximum to be robust against outliers.

We also include an additional feature per axis, lag-1 autocorrelation. Autocorrelation measures the correlation between a signal's current value and its past values. By taking a lag of 1, we are considering  how correlated the values of the signal are with their previous value.

In total, each window has five percentile features per axis and one lag-1 correlation feature per axis, for a total of 18 features per window.

Each activity bout in the dataset consists of a number of minutes of activity. For use with our method, each minute of activity is broken up into five 12-second windows. This length was chosen because it was the lowest reasonable limit we could justify to maximize the resolution of our time series signals. The fewer windows the activity bout has, the coarser the cluster summary features would be.

However, it must be addressed that 12-second intervals do not evenly divide into 10th, 25th, 50th, 75th, and 90th percentiles. For this, we chose to use the nearest appropriate points for those percentiles: the 2nd, 3rd, 6th, 9th, and 11th points. Since our data resolution was 1-second, we are very limited in our choices of window-lengths. The smallest evenly-divisible length would be 20. However, we chose 12 to grant us additional resolution beyond that limit. An alternative to choosing the nearest appropriate points would be to perform interpolation, and in this setting we cannot justify interpolating the data at these scales. Since we cannot account for the general behavior of the signal between measurements, there is no reasonable assumption to make for a best approximation of in-between points; choosing any interpolation scheme would introduce significant bias.

All regression, ANN, and experiments performed herein were done using MATLAB 2018a \cite{MATLAB:2018}. The code for Variational Bayesian Gaussian Mixture Models is part of the Pattern Recognition and Machine Learning Toolbox on the MATLAB File Exchange \cite{VBGMmatlab}.

\subsection{Leave-one Person Out Cross Validation}

For cross validating our method, we use leave-one person out. We train our framework on all of the data excluding the instances associated with a single participant, and repeat for each participant in the data set. We chose to test at the participant scope rather than the instance scope to avoid any bias we would incur while predicting PA type or estimating PAEE for one of a participant's activities given that we have trained on the other activities they performed for the dataset. 

\subsection{Competitive Methods}

For the classification phase, we compare \textit{SummerTime} with a baseline Artificial Neural Network (ANN) model.

For the regression phase, we compare \textit{SummerTime} with Linear Regression on Local Features with Voting, on a 5-Regression Model applied to the classification by ANN, and to ANN on Local Features with Voting.

\subsubsection{ANN Classification with Voting}

The ANN we compare classification against takes as input the time window features and produces an activity type classification per window. The classification result for the entire activity is aggregated by majority voting, using the model. The hidden layer consists of 25 hidden neurons, following the network topology from \cite{Staudenmayer2009}.

Comparison with this method shows that the summarization features produce better results than using the local features with voting, since the underlying classification method is equivalent.

\subsubsection{Linear Regression on Local Features}

We compare the regression phase of \textit{SummerTime} with a linear regression model. The design matrix of the linear regression consists of a bias term and the window features. The PAEE estimates are aggregated as sums of each window for the activity as a whole. 

This model is a regression baseline for the bare minimum attempt at estimating the energy expenditure from available variables.

\subsubsection{5-regression splitting on ANN classification}

We also compare the regression phase with the 5-regression model but using the classification result of the ANN classifier. The regression model does not include the summarization features as input variables, but is otherwise the same as described in Section 3.3. Again,  regression results are aggregated as sums for the activity as a whole.

Comparison with this method shows that our summarization features provide a notable improvement over performing the 5-regression regression alone.

\subsubsection{ANN Regression}

Finally, we compare the regression phase with the ANN directly. The ANN is a powerful method that is one of the main methods used for classifying physical activity type. Once again, regression results are aggregated as sums for the activity as a whole.

\begin{table*}
    \centering
        \begin{tabulary}{1\textwidth}{|c|c|c|c|c|c|}
        \hline          & \bfseries Sed. & \bfseries LHH & \bfseries MtV & \bfseries Walk & \bfseries Run \\
        \hline \bfseries Sed.  &    \textbf{16455} &	20	& 0	& 0	& 0 \\
        \hline \bfseries LHH     &    90	& \textbf{2085}	& 310	& 20	&  0  \\
        \hline \bfseries MtV     &    20	& 240 & 	\textbf{1290} & 	20	 & 0 \\
        \hline \bfseries Walk  &    25	& 0	  &  25	  &   \textbf{3685}	 & 40 \\
        \hline \bfseries Run &    0	& 0	   & 0	    &     135	 & \textbf{350} \\
        \hline 
        \end{tabulary}
    \caption{Above, we have the counts of the Confusion Matrix of activity instances for the Classification Phase. Columns are predictions and rows are actuals. Along the diagonal are the in-class true-counts.}
    \label{tab:confusion_matrix1}
\end{table*}

\begin{table*}
    \centering
        \begin{tabulary}{1\textwidth}{|c|c|c|c|c|c|}
                 \hline
                 & Sed.  & LHH   & MtV   & Walk  & Run   \\ \hline
            Sed. & \bfseries 99.88 &  0.12 &  0.00 &  0.00 &  0.00 \\ \hline
            LHH	 &  3.59 & \bfseries 83.23 & 12.38 &  0.80 &  0.00 \\ \hline
            MtV	 &  1.27 & 15.29 & \bfseries 82.17 &  1.27 &  0.00 \\ \hline
            Walk &  0.66 &  0.00 &  0.66 & \bfseries 97.62 &  1.06 \\ \hline
            Run	 &  0.00 &  0.00 &  0.00 & 27.84 & \bfseries 72.16 \\ \hline
        \end{tabulary}
    \caption{Here, we show the Confusion Matrix for the Classification Phase as percentages, with Recall along the diagonal. Columns are predictions and rows are actuals.}
    \label{tab:confusion_matrix2}
\end{table*}

\subsection{Classification Results}

In validating PA classification, we use multi-class accuracy. We will generally refer to in-class accuracy, as opposed to overall accuracy. The in-class accuracies of each class are more indicative of the success of our method as they are not heavily influenced by the large number of easily-correctly-classified Sedentary activities. The changes in overall accuracy are less indicative of improvement because of the imbalance in the underlying dataset.

In Table \ref{tab:confusion_matrix1}, we show a confusion matrix detailing the classification of activity instances. Rows represent the ground truth labels of each instance, whereas columns represent our classification model's predictions. Along the diagonals in bold script are the correctly classified instances. Along the diagonal of Table \ref{tab:confusion_matrix2}, we see the recall of the classifier.


\begin{table*}
    \centering
        \begin{tabulary}{1\textwidth}{|c|c|c|c|c|c|}
                 \hline
     & Sed.  & LHH   & MtV   & Walk  & Run                \\ \hline
Sed. & \bfseries 99.64 &	 0.36 &	 0.00 &	 0.00 &	 0.00 \\ \hline
LHH	 &  0.00 &	\bfseries 87.82 &	 6.59 &	 3.99 &	 1.60 \\ \hline
MtV	 &  0.00 &	68.15 &	\bfseries 31.85 &	 0.00 &	 0.00 \\ \hline
Walk &  0.66 &	 4.24 &	 0.00 &	\bfseries 85.56 &	 9.54 \\ \hline
Run	 &  0.00 &	16.49 &	 0.00 &	75.26 &	\bfseries    8.25 \\ \hline
\end{tabulary}
    \caption{Here, we show the Confusion Matrix for the ANN classification as percentages, with Recall along the diagonal. Columns are predictions and rows are actuals. Take note of the misclassification of Run activities as Walking and Light Household Chores and Games.}
    \label{tab:confusion_matrix3}
\end{table*}

In Table \ref{tab:confusion_matrix3}, we show the classification results of an ANN, which achieves poor Run category recall. A recall of 8.25\% is far worse than of random guessing (an expectation of 20\% for a 5-class classification). This indicates that the model is biased against making predictions of Running. While the ANN has a higher recall for the Light Household Chores category, it is misclassifying a larger portion of instances outside of the LHH category.

\subsection{Regression Results}

\begin{table*}
    \centering
    \begin{tabulary}{1\textwidth}{| l | l | l | l | l | l || l |}
        \hline
                      & \bfseries Sed. & \bfseries LHH & \bfseries MtV & \bfseries Walk & \bfseries Run & \bfseries Overall \\ \hline
            \bfseries       Linear Regression       & 2.0105 & 2.7549 & 3.3990 & 2.6670 & 4.1593 & 2.3690 \\ \hline
            \bfseries       5-Regression  & 0.1787 & 1.2631 & 1.6737 & 1.3500 & 1.7201 & 0.8346 \\ \hline
            \bfseries ANN & 0.3999 & 2.2715 & 2.7789 & 2.1308 & 3.6695 & 1.4402 \\ \hline
            \bfseries Our Method & \bfseries 0.1741 & \bfseries 1.0406 & \bfseries 1.4231 & \bfseries 1.2268 & \bfseries 1.2693 & \bfseries 0.7206 \\ \hline
            
    \end{tabulary}
    \caption{Above, we show the RMSE for MET estimates of each Regression Model by Activity Class. Our method performs best in each class and overall. Note the similarity of RMSEs between Walking and Running, a similarity not experienced by any of the other methods. This indicates that the intensity of the activity doesn't have an impact on precision of EE estimates.}
    \label{tab:rmse1}
\end{table*}

In validating PAEE regression results, we make use of the root mean-squared error (RMSE) measure. The formula for RMSE is as follows, where $\hat y$ is the regression output and $y$ is the true EE (Energy Expenditure).

\begin{equation}
    \text{RMSE} = \sqrt{\mathbb{E}\left(\left(\hat y - y\right)^2\right)}
\end{equation}

This measure acts as the sample standard deviation of differences between predicted and true values, meaning that we can interpret it directly as a measure of both accuracy and precision of prediction \cite{wiki:RMSE}. We calculate RMSE for both in-class and overall.

In Table \ref{tab:rmse1}, we show the RMSE values associated with the predicted METs for each regression model tested by every activity category. Our method out-performed each method overall and in each individual activity category as well, achieving an overall RMSE of 0.7206. It should be noted that the RMSE associated with the Running is much larger than that of the Walking category in other regression methods. In our method's predictions, the RMSE values of both categories are much closer together, differing by only 0.0425 units. This indicates that the intensity of the activity doesn't have an impact on precision of EE estimates in our method.

\section{Related Work}
Here we present a list of major works that have been done on physical activity classification and energy expenditure predictions using machine learning methods. We also include important works in the greater field of time series analysis.

In 2005, Crouter et al. introduce the two-regression model which alternates between a quadratic regression model and a linear regression model based on the coefficients of variations of each bout. \cite{Crouter1324} This novel approach broke the overall problem objective into two key parts: first, separating instances by their variability into two groupings based on their coefficients of variation; second, applying to each grouping a regression model which is more appropriate for instances of that variability. 

In 2009, Ye et al. develop a new time series primitive called a shapelet. The shapelet is a subsequence pattern intended to be maximally representative of some class of time series. They demonstrate in the work that the new concept is time-efficient, accurate, and interpretable \cite{ye2009time}.

In 2012, Trost et al. was able to improve physical activity classification accuracy as well as low root mean squared-error (RMSE) in energy expenditure estimation with an Artificial Neural Network (ANN) model. \cite{Trost}

In that same year, Mu et al., revisited the two-regression model of Crouter et al. and extended it to a number of regression models, one per each activity type \cite{TKDE}. The data used in this study including each activity bout therein was structured rather variably, which made it analogous to a free-living data collection. This method utilized distance metric learning methods to learn the underlying block structure of variable-length activity bouts. 

In 2014, Staudenmayer et al. expanded on the field with another ANN model which they applied to their own dataset \cite{Staudenmayer2009}. However, their classification procedure was targeting learned activity types, as opposed to expert-defined types. They produced these types through clustering based on their signal activity levels.

Petitjean et al. developed a nearest centroid classification method which constructs centroids that are meaningful to the Dynamic Time Warping (DTW) Algorithm, to allow for time-efficient classification with the distance-based approach \cite{petitjean2014dynamic}.

Bastion et al. published an evaluation of cutting-edge methods outside of the rigid laboratory setting and confirmed the activity classification community's suspicions that existing methods would not perform well in the free-living setting \cite{Bastian716}.

Hills et al. developed a time series classification method using shapelets to produce an alternative representation of time series signals where the new features are distances from each of $k$ shapelets \cite{hills2014classification}.

In 2015, Baydogan et al. made use of the dependency structure within time series to develop a representation and similarity measure which they validate on a wide variety of time series domains \cite{baydogan2016time}.

In 2017, David Hallac et al. developed a method that segments and clusters time series data using structural network relations rather than spatial distance to encode different groupings of time series segments \cite{Hallac:2017:TIC:3097983.3098060}.

In 2018, Stanislas Chambon et al. developed a deep learning approach for sleep stage classification that learns end-to-end without computing spectrograms or extracting handcrafted features \cite{chambon2018deep}.

In 2019, Fazle Karim et al. proposed transforming the existing univariate time series classification models, LSTM-FCN and ALSTM-FCN, into a multivariate time series classification model by augmenting the fully convolutional block with a squeeze-and-excitation block to further improve accuracy \cite{karim2019multivariate}.

\section{Discussion}
To resolve the lack of classical structure in the time series data, in this paper, we chose to bridge the gap between the variable-length, empirically-chosen features for physical activity data and a latently-learned fixed-length representation, or summary. Our method, \textit{SummerTime}, leverages an existing feature construction and through clustering of disjoint time windows establishes a summary of the time series as a whole. The features of the summarization were extracted from the data unsupervised through Gaussian mixture using a variational Bayesian approach. This allows our method to zero-in on the number of representative features of the summarization, naturally doing feature selection. This clustering provides us with an informative fixed-length feature-vector which contains a global description of the signal for a single activity bout. From this, we are able to preform classical machine learning methods on the summaries instead of the original instances. We show that this summarization is actually sufficient for problems like physical activity type classification and out-performs classification on each time window independently with majority voting. We then show that \textit{SummerTime} can augment energy expenditure predictions per window by including with each window's original features the summarization of the bout. In what we believe to be a spectacular achievement, \textit{SummerTime} manages to get small, comparable error for both Walking and Running type activities, two activity types which normally have significant cross-class error. Overall, \textit{SummerTime} demonstrates low classification and regression error overall, robustness, and effectiveness in an imbalanced dataset. We hope to further demonstrate the strengths of \textit{SummerTime} in more time series domains in the future.

\section{Acknowledgements}

[Content removed in compliance with Triple Blind. Will be added back in after Triple Blind period.]

\bibliography{main}{}

\begin{thebibliography}{10}

\bibitem{Bastian716}
Thomas Bastian, Aur{\'e}lia Maire, Julien Dugas, Abbas Ataya, Cl{\'e}ment
  Villars, Florence Gris, Emilie Perrin, Yanis Caritu, Maeva Doron,
  St{\'e}phane Blanc, Pierre Jallon, and Chantal Simon.
\newblock Automatic identification of physical activity types and sedentary
  behaviors from triaxial accelerometer: laboratory-based calibrations are not
  enough.
\newblock {\em Journal of Applied Physiology}, 118(6):716--722, 2015.

\bibitem{baydogan2016time}
Mustafa~Gokce Baydogan and George Runger.
\newblock Time series representation and similarity based on local
  autopatterns.
\newblock {\em Data Mining and Knowledge Discovery}, 30(2):476--509, 2016.

\bibitem{bishop2006pattern}
Christopher~M Bishop.
\newblock Mixture models and em.
\newblock In {\em Pattern recognition and machine learning}, chapter~9.
  Springer, 2006.

\bibitem{bishop2006pattern5}
Christopher~M Bishop.
\newblock Neural networks.
\newblock In {\em Pattern recognition and machine learning}, chapter~5.
  Springer, 2006.

\bibitem{chambon2018deep}
Stanislas Chambon, Mathieu~N Galtier, Pierrick~J Arnal, Gilles Wainrib, and
  Alexandre Gramfort.
\newblock A deep learning architecture for temporal sleep stage classification
  using multivariate and multimodal time series.
\newblock {\em IEEE Transactions on Neural Systems and Rehabilitation
  Engineering}, 26(4):758--769, 2018.

\bibitem{Crouter1324}
Scott~E. Crouter, Kurt~G. Clowers, and David~R. Bassett.
\newblock A novel method for using accelerometer data to predict energy
  expenditure.
\newblock {\em Journal of Applied Physiology}, 100(4):1324--1331, 2006.

\bibitem{freedson2011evaluation}
Patty~S Freedson, Kate Lyden, Sarah Kozey-Keadle, and John Staudenmayer.
\newblock Evaluation of artificial neural network algorithms for predicting
  mets and activity type from accelerometer data: validation on an independent
  sample.
\newblock {\em Journal of Applied Physiology}, 111(6):1804--1812, 2011.

\bibitem{Hallac:2017:TIC:3097983.3098060}
David Hallac, Sagar Vare, Stephen Boyd, and Jure Leskovec.
\newblock Toeplitz inverse covariance-based clustering of multivariate time
  series data.
\newblock In {\em Proceedings of the 23rd ACM SIGKDD International Conference
  on Knowledge Discovery and Data Mining}, KDD '17, pages 215--223, New York,
  NY, USA, 2017. ACM.

\bibitem{hills2014classification}
Jon Hills, Jason Lines, Edgaras Baranauskas, James Mapp, and Anthony Bagnall.
\newblock Classification of time series by shapelet transformation.
\newblock {\em Data Mining and Knowledge Discovery}, 28(4):851--881, 2014.

\bibitem{karim2019multivariate}
Fazle Karim, Somshubra Majumdar, Houshang Darabi, and Samuel Harford.
\newblock Multivariate lstm-fcns for time series classification.
\newblock {\em Neural Networks}, 116:237--245, 2019.

\bibitem{MATLAB:2018}
MATLAB.
\newblock {\em version R2018a}.
\newblock The MathWorks Inc., Natick, Massachusetts, 2018.

\bibitem{TKDE}
Y.~Mu, H.~Z. Lo, W.~Ding, K.~Amaral, and S.~E. Crouter.
\newblock Bipart: Learning block structure for activity detection.
\newblock {\em IEEE Transactions on Knowledge and Data Engineering},
  26(10):2397--2409, Oct 2014.

\bibitem{nasios2006variational}
Nikolaos Nasios and Adrian~G Bors.
\newblock Variational learning for gaussian mixture models.
\newblock {\em IEEE Transactions on Systems, Man, and Cybernetics, Part B
  (Cybernetics)}, 36(4):849--862, 2006.

\bibitem{VBGMmatlab}
M. {Chen}, {Pattern Recognition and Machine Learning Toolbox}, 2016.
\newblock {MATLAB} File Exchange.

\bibitem{petitjean2014dynamic}
Fran{\c{c}}ois Petitjean, Germain Forestier, Geoffrey~I Webb, Ann~E Nicholson,
  Yanping Chen, and Eamonn Keogh.
\newblock Dynamic time warping averaging of time series allows faster and more
  accurate classification.
\newblock In {\em Data Mining (ICDM), 2014 IEEE International Conference on},
  pages 470--479. IEEE, 2014.

\bibitem{Staudenmayer2009}
J.~Staudenmayer, D.~Pober, S.~Crouter, D.~Bassett, and P.~Freedson.
\newblock {{A}n artificial neural network to estimate physical activity energy
  expenditure and identify physical activity type from an accelerometer}.
\newblock {\em J. Appl. Physiol.}, 107(4):1300--1307, Oct 2009.

\bibitem{Trost}
S.~G. Trost, W.~K. Wong, K.~A. Pfeiffer, and Y.~Zheng.
\newblock {{A}rtificial neural networks to predict activity type and energy
  expenditure in youth}.
\newblock {\em Med Sci Sports Exerc}, 44(9):1801--1809, Sep 2012.

\bibitem{wiki:RMSE}
{Wikipedia contributors}.
\newblock Root-mean-square deviation --- {Wikipedia}{,} the free encyclopedia,
  2018.

\bibitem{ye2009time}
Lexiang Ye and Eamonn Keogh.
\newblock Time series shapelets: a new primitive for data mining.
\newblock In {\em Proceedings of the 15th ACM SIGKDD international conference
  on Knowledge discovery and data mining}, pages 947--956. ACM, 2009.

\end{thebibliography}
\bibliographystyle{plain}

\end{document}